\documentclass[11pt,a4paper]{article}
\usepackage[hyperref]{emnlp2020}
\usepackage{times}
\usepackage{latexsym}
\usepackage{graphicx}
\usepackage{booktabs}
\usepackage{placeins}
\usepackage{multirow, makecell}
\usepackage{amsmath,amssymb}

\usepackage{enumitem}

\usepackage{microtype}

\aclfinalcopy 
\def\aclpaperid{1971} 

\newcommand\comment[1]{}
\newcommand\elmo{\textsc{ELMo}}
\newcommand\bert{\textsc{BERT}}
\newcommand\seqtoseq{\textsc{Seq2seq}}
\newcommand\grammar{\textsc{Grammar}}
\newcommand\attnsup{\textsc{AttnSup}}
\newcommand\attnspan{\textsc{AttnSpan}}
\newcommand\coverage{\textsc{Coverage}}
\newcommand{\emprog}{\text{EM}_{\text{program}}}
\newcommand{\emiid}{\text{EM}_{\text{iid}}}

\newcommand\atis{\textsc{ATIS}}
\newcommand\geo{\textsc{GeoQuery}}
\newcommand\scholar{\textsc{Scholar}}
\newcommand\advising{\textsc{Advising}}
\newcommand\drop{\textsc{DROP}}

\newcommand\nl[1]{\emph{``#1''}}

\newif\ifcomments
\ifcomments
    \providecommand\jb[1]{\textcolor{red}{[JB: {#1}]}}
    \providecommand\jh[1]{\textcolor{blue}{[JH: {#1}]}}
    \providecommand\io[1]{\textcolor{orange}{[IO: {#1}]}}
    \providecommand\nitish[1]{\textcolor{teal}{[NG: {#1}]}}
    \providecommand\matt[1]{\textcolor{purple}{\bf [Matt: {#1}]}}
    \providecommand\todo[1]{\textcolor{red}{\bf [TODO: {#1}]}}
\else
    \providecommand\jb[1]{}
    \providecommand\jh[1]{}
    \providecommand\io[1]{}
    \providecommand\nitish[1]{}
    \providecommand\matt[1]{}
    \providecommand\todo[1]{}
\fi

\title{Improving Compositional Generalization in Semantic Parsing}
\author{Inbar Oren$^{1}$ ~~~ Jonathan Herzig\thanks{\;\;\; The authors contributed equally.}$^{*,1}$ ~~~ Nitish Gupta$^{*,2}$ ~~~ Matt Gardner$^{3}$ ~~~
Jonathan Berant$^{1,3}$ \\
\mbox{}\\
$^1$School of Computer Science, Tel-Aviv University \\
$^2$University of Pennsylvania \\
$^3$Allen Institute for AI \\
\small{\texttt{\{inbaroren@mail, jonathan.herzig@cs, joberant@cs\}.tau.ac.il,}} \\
\small{\texttt{nitishg@seas.upenn.edu, mattg@allenai.org}}
}

\date{}

\begin{document}
\maketitle
\begin{abstract}
Generalization of models to out-of-distribution (OOD) data has captured tremendous attention recently. Specifically, \emph{compositional generalization},
i.e., whether a model generalizes to new structures built of components observed during training, has sparked substantial interest. In this work, 
we investigate compositional generalization in semantic parsing, a natural test-bed for compositional generalization, as output programs are constructed from sub-components.
We analyze a wide variety of models and propose multiple extensions to the attention module of the semantic parser, aiming to improve compositional generalization.
We find that the following factors improve compositional generalization: (a) using contextual representations, such as \elmo{} and \bert{},
(b) informing the decoder what input tokens have previously been attended to,
(c)  training the decoder attention to agree with
pre-computed token alignments,
and (d) 
downsampling examples corresponding to frequent program templates. 
While we substantially reduce the gap between in-distribution and OOD generalization, performance on OOD compositions is still substantially lower.
 
\end{abstract}

\section{Introduction}

Neural models trained on large datasets have recently shown great performance on data sampled from the training distribution. However, generalization to out-of-distribution (OOD) scenarios has been dramatically lower \cite{sagawa2019distributionally, gardner2020evaluating, kaushik2019learning}. A particularly interesting case of OOD generalization is \emph{compositional generalization}, the ability to systematically generalize to test examples composed of components seen during training. For example, we expect a model that observed the questions \nl{What is the capital of France?} and \nl{What is the population of Spain?} at training time to generalize to questions such as \nl{What is the population of the capital of Spain?}.
While humans generalize systematically to such compositions \cite{fodor1988connectionism}, models often fail to capture the
structure underlying the problem, and thus miserably fail \cite{atzmon2016learning, lake-2018-scan, loula-etal-2018-rearranging, Bahdanau2019CLOSUREAS,ruis2020benchmark}. 

Semantic parsing, mapping natural language utterances to structured programs, is a task where compositional generalization is expected, as sub-structures in the input utterance and output program often align. For example, in \nl{What is the capital of the largest US state?}, the span \nl{largest US state} might correspond to an \texttt{argmax} clause in the output program.
Nevertheless, prior work \cite{finegan-dollak-etal-2018-improving, herzig2019detect, keysers2020measuring} has shown that data splits that require generalizing to new program templates result in drastic loss of performance.
However, past 
work did not investigate how different modeling choices interact with compositional generalization.

In this paper, we thoroughly analyze the impact of different modeling choices on compositional generalization in 5 semantic parsing datasets---four that are text-to-SQL datasets, and \drop{}, a dataset for executing programs over text paragraphs.
Following \newcite{finegan-dollak-etal-2018-improving}, we examine performance on a \emph{compositional split}, where target programs are partitioned into ``program templates", and templates appearing at test time are unobserved at training time. We examine the effect of standard practices, such as contextualized representations~(\S\ref{ssec:model-repr}) and grammar-based decoding~(\S\ref{ssec:model-grammar}). Moreover, we propose novel extensions to decoder attention~(\S\ref{ssec:model-attn}), the component responsible for aligning sub-structures in the question and program: (a) supervising attention based on pre-computed token alignments, (b) attending over constituency spans, and (c) encouraging the decoder attention to cover the entire input utterance. Lastly, we also propose downsampling examples from frequent templates to reduce dataset bias~(\S\ref{ssec:model-skew}).

Our main findings are that (i) contextualized representations, (ii) supervising the decoder attention, (iii) informing the decoder on coverage of the input by the attention mechanism, and (iv) downsampling frequent program templates, all reduce the gap in generalization when comparing standard iid splits to compositional splits. For SQL, the gap in exact match accuracy between in-distribution and OOD is reduced from $84.6 \rightarrow 62.2$ and for \drop{} from $96.4 \rightarrow 77.1$.
While this is a substantial improvement, the gap between in-distribution and OOD generalization is still significant.
All our code and data are publicly available at \url{http://github.com/inbaroren/improving-compgen-in-semparse}.
\section{Compositional Generalization}
\label{sec:comp_gen}
Natural language is compositional in a sense that complex utterances are interpreted by understanding the structure of the utterance and the meaning of its parts~\cite{Montague1973}. For example, the meaning of \nl{a person below the tree} can be composed from the meaning of \nl{a person}, \nl{below} and \nl{tree}. By virtue of compositionality, an agent can derive the meaning of new utterances, even at first encounter.  
Thus, we expect our systems to model this compositional nature of language and generalize to new utterances, generated from sub-parts observed during training but composed in novel ways. This sort of model generalization is often called \emph{compositional generalization}. 

Recent work has proposed various benchmarks to measure different aspects of \emph{compositional generalization}, showing that current models struggle in this setup.
\citet{lake-2018-scan} introduce a benchmark called SCAN for mapping a command to actions in a synthetic language, and proposed a data split that requires generalizing to commands that map to a longer sequence of actions than observed during training.
\citet{Bahdanau2019SystematicGW} study the impact of modularity in neural models on the ability to answer visual questions about pairs of objects that were not observed during training. \citet{Bahdanau2019CLOSUREAS} assess the ability of models trained on CLEVR~\cite{clevr-2017-johnson} to interpret new referring expressions composed of parts observed at training time. \citet{keysers2020measuring} develop a benchmark of Freebase questions and propose a data split such that the test set contains new combinations of knowledge-base constants (entities and relations) that were not seen during training. \newcite{ruis2020benchmark} proposed gSCAN, which focuses on compositional generalization when mapping commands to actions in a situated environment.
 
\begin{figure}[t]
  \includegraphics[width=1.0\columnwidth]{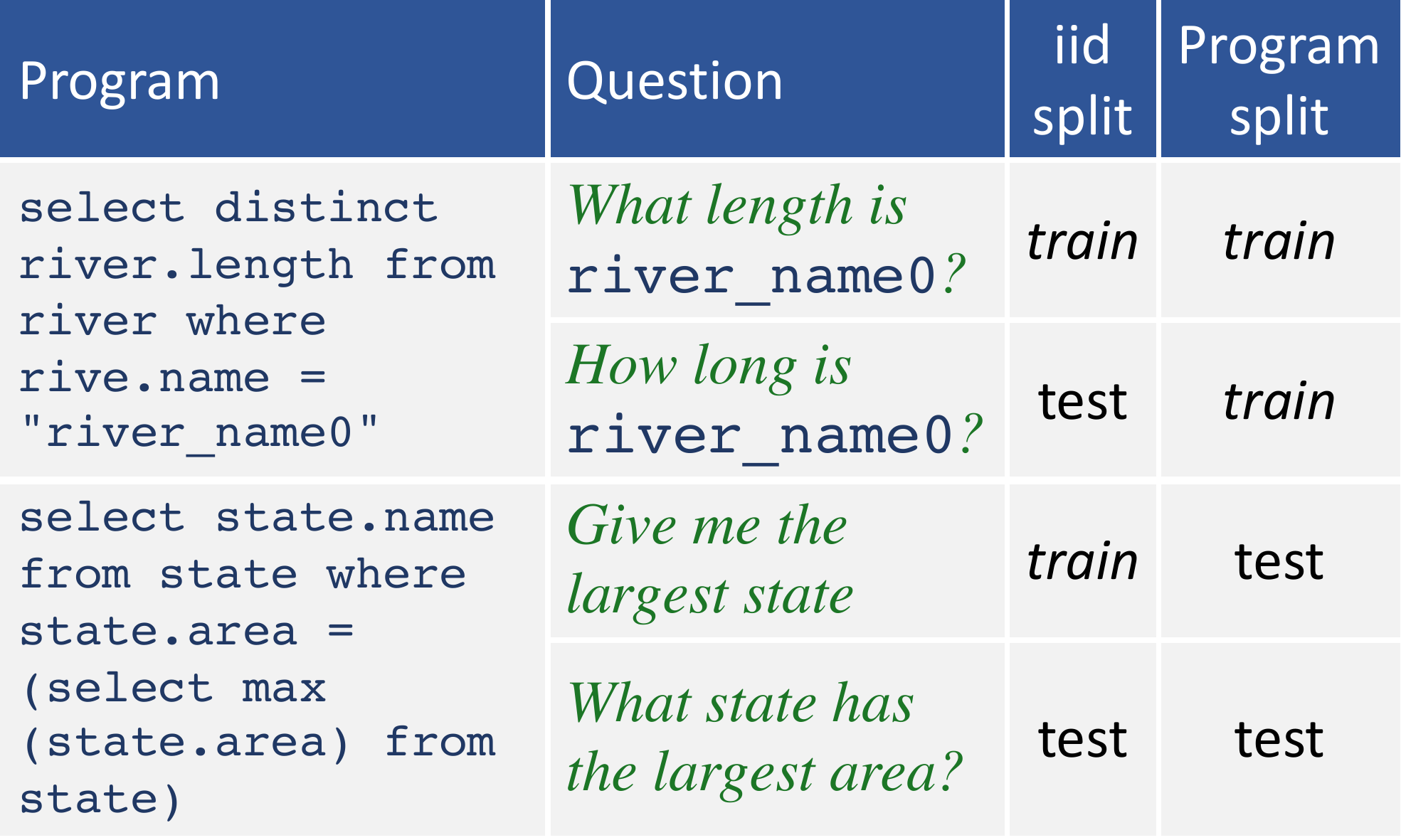}
  \caption{An iid split of examples in semantic parsing leads to identical anonymized programs appearing at both training and test time. A program split prohibits anonymized programs from appearing in the same partition, and hence tests compositional generalization.
  }
  \label{fig:example_split}
\end{figure}

In this work, we focus on a specific kind of compositional data split, proposed by \citet{finegan-dollak-etal-2018-improving}, that despite its simplicity leads to large drops in performance.
\citet{finegan-dollak-etal-2018-improving} propose to split semantic parsing data such that a model cannot memorize a mapping from question templates to programs. 
To achieve this, they take question-program pairs, and \emph{anonymize} the entities in the question-program pair with typed variables. Thus, questions that require the same abstract reasoning structure now get mapped to the same anonymized program, referred to as \emph{program template}. 
For example, in the top two rows of Figure~\ref{fig:example_split}, after anonymizing the name of a river to the typed variable \texttt{river\_name0}, two lexically-different questions map to the same program template. Similarly, in the bottom two rows we see two different questions that map to the same program even before anonymization.

The data is then split in a manner such that a program template and all its accompanying questions belong to the same set, called the \emph{program split}. This ensures that all test-set program templates are unobserved during training. 
For example, in a \emph{iid split} of the data, it is possible that the question \nl{what is the capital of France?} will appear in the training set, and the question \nl{Name Spain's capital.} will appear in the test set. Thus, the model only needs to memorize a mapping from question templates to program templates.
However, in the \emph{program split}, each program template is in either the training set or test set, and thus a model must generalize at test time to new combinations of predicates and entities~(see Figure~\ref{fig:example_split} - Program split).

We perform the compositional split proposed by \newcite{finegan-dollak-etal-2018-improving} on four text-to-SQL datasets from \newcite{finegan-dollak-etal-2018-improving} and one dataset for mapping questions to QDMR programs~\cite{break2020} on DROP~\cite{drop2019}. Exact experimental details are in \S\ref{sec:datasets}.

\section{Model}
\label{sec:model}
\newcite{finegan-dollak-etal-2018-improving} convincingly showed that a program split leads to low semantic parsing performance. However, they examined only a simple baseline parser, disregarding many standard variations that have been shown to improve in-distribution generalization, and might affect OOD generalization as well. In this section, we describe variants to both the model and training, and evaluate their effect on generalization in \S\ref{sec:experiments}. We examine well-known choices, such as the effect of contextualized representations (\S\ref{ssec:model-repr}) and grammar-based decoding (\S\ref{ssec:model-grammar}), as well as several novel extensions to the decoder attention (\S\ref{ssec:model-attn}), which include (a) eliciting supervision (automatically) for the decoder attention distribution, (b) allowing attention over question spans, and (c) encouraging attention to cover all of the question tokens. For \drop{}, where the distribution over program templates is skewed, we also examine the effect of reducing this bias by downsampling frequent program templates (\S\ref{ssec:model-skew}).

\comment{
Over the years, the community has proposed different modeling variations that affect the in-distribution performance of semantic parsers. 
In this work we measure the impact of commonly used modeling choices on compositional generalization in semantic parsers. Specifically, we measure the effect of contextualized representations~(\S\ref{ssec:model-repr}) and grammar-based decoding~(\S\ref{ssec:model-grammar}) on parsing accuracy.
We also propose novel modeling extensions to decoder attention and examine their effect on compositional generalization: (a) supervising decoder attention based on pre-computed token alignments, (b) attending over question spans from a constituency parser, and (c) encouraging high attention coverage as performed in text summarization models~\cite{see2017get} (\S\ref{ssec:model-attn}). 
Additionally, we propose downsampling training examples from frequent templates to reduce skewness of the training data~(\S\ref{ssec:model-skew}).
}

\paragraph{Baseline Semantic Parser} 

A semantic parser maps an input question $x$ into a program $z$, and in the supervised setup is trained from $(x,z)$ pairs.
Similar to \newcite{finegan-dollak-etal-2018-improving}, our baseline semantic parser is a standard sequence-to-sequence model~\cite{dong-lapata-2016-language} that encodes the question $x$ with a BiLSTM encoder \cite{hochreiter1997long} over GloVe embeddings~\cite{pennington2014glove}, and decodes the program $z$ token-by-token from left to right with an attention-based LSTM decoder~\cite{bahdanau2015neural}.

\subsection{Contextualized Representations}
\label{ssec:model-repr}
Pre-trained contextualized representations revolutionized natural language processing in recent years, and semantic parsing has been no exception \cite{guo-etal-2019-towards, wang2019rat}.
We hypothesize that better representations for question tokens should improve compositional generalization, because they reduce language variability and thus may help improve the mapping from input to output tokens.
We evaluate the effect of using \elmo{} \cite{peters2018elmo} and \bert{} \cite{devlin-etal-2019-bert} 
to represent question tokens.\footnote{We use fixed \bert{} embeddings without fine-tuning in the SQL datasets due to computational constraints.}

\subsection{Grammar-Based Decoding}
\label{ssec:model-grammar}
A unique property of semantic parsing, compared to other generation tasks, is that programs have a clear hierarchical structure that is based on the target formal language. Decoding the output program token-by-token from left to right \cite{dong-lapata-2016-language, jia2016recombination} can thus generate programs that are not syntactically valid, and the model must effectively learn the syntax of the target language at training time. Grammar-based decoding resolves this issue and has been shown to consistently improve in-distribution performance \cite{rabinovich-etal-2017-abstract, krishnamurthy2017neural, yin-neubig-2017-syntactic}.
In grammar-based decoding, the decoder outputs the abstract syntax tree of the program based on a formal grammar of the target language. At each step, a production rule from the grammar is chosen, eventually outputting a top-down left-to-right linearization of the program tree. Because decoding is constrained by the grammar, the model outputs only valid programs. 
We refer the reader to the aforementioned papers for details on grammar-based decoding.

\comment{
\begin{figure}
  \centering
  \includegraphics[width=\columnwidth]{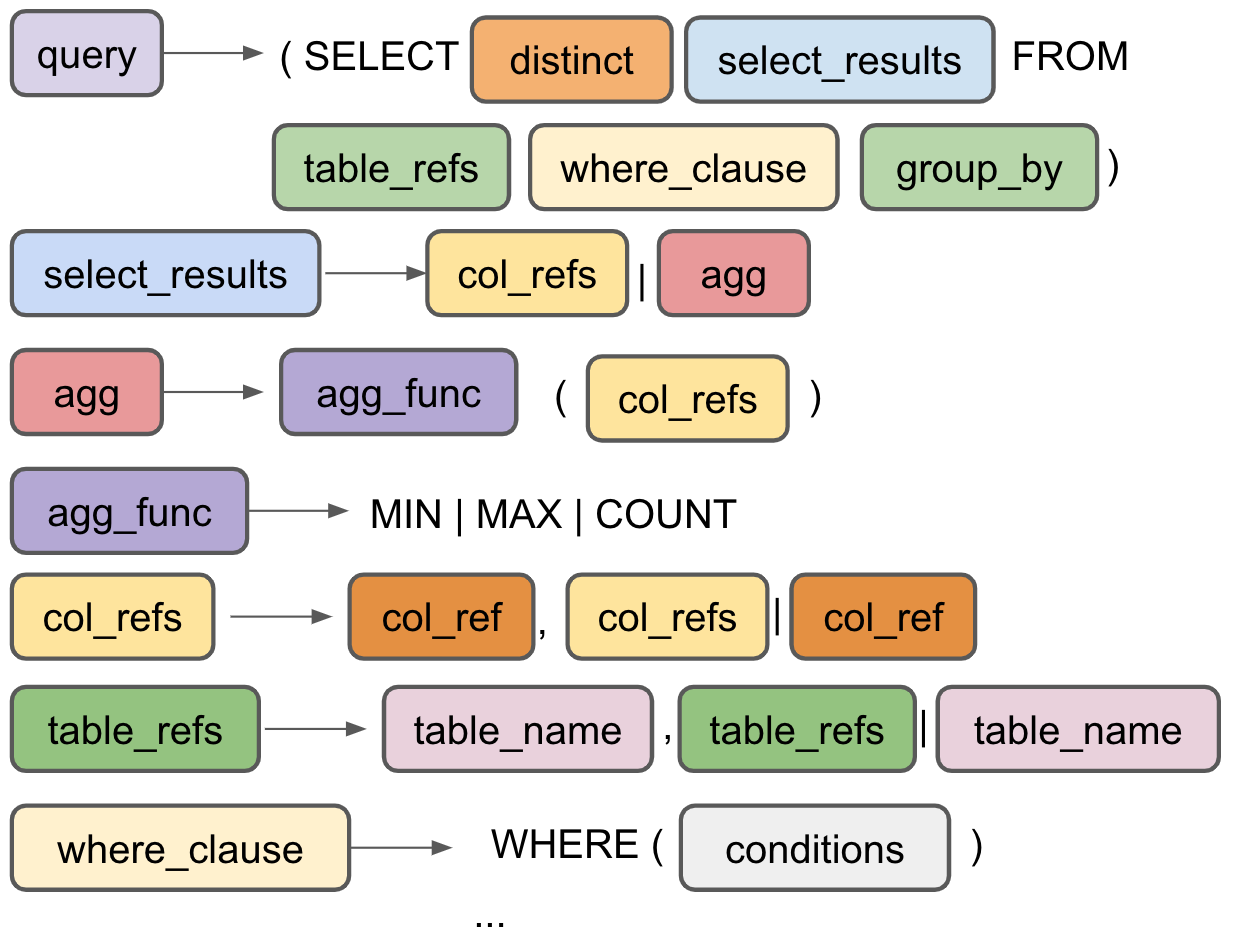}
  \caption{A sketch of the rules of a grammar for SQL.
  }
  \label{fig:grammar}
\end{figure} 
}

Compositional generalization involves combining known sub-structures in novel ways. In grammar-based decoding, the structure of the output program is explicitly generated, and this could potentially help compositional generalization.
We discuss the grammars used in this work in \S\ref{sec:datasets}.

\subsection{Decoder Attention}
\label{ssec:model-attn}

Semantic parsers use attention-based decoding:
at every decoding step, the model computes a distribution $(p_1 \dots p_n)$ over the question tokens $x = (x_1, \dots, x_n)$ and the decoder computes its next prediction based on the weighted average $\sum_{i=1}^n p_i \cdot h_i$, where $h_i$ is the encoder representation of $x_i$. Attention has been shown to both improve in-distribution performance \cite{dong-lapata-2016-language} and also lead to better compositional generalization~\cite{finegan-dollak-etal-2018-improving}, by learning a soft alignment between question and program tokens.
Since attention is the component in a sequence-to-sequence model that aligns parts of the input to parts of the output, we propose new extensions to the attention mechanism, and examine their effect on compositional generalization.

\paragraph{(a) Attention Supervision} 
Intuitively, learning good alignments between question and program tokens should improve compositional generalization: a model that correctly aligns the token \emph{largest} to the predicate \texttt{max} should output this predicate when encountering \emph{largest} in novel contexts.

\begin{figure}[t]
  \includegraphics[width=0.5\textwidth]{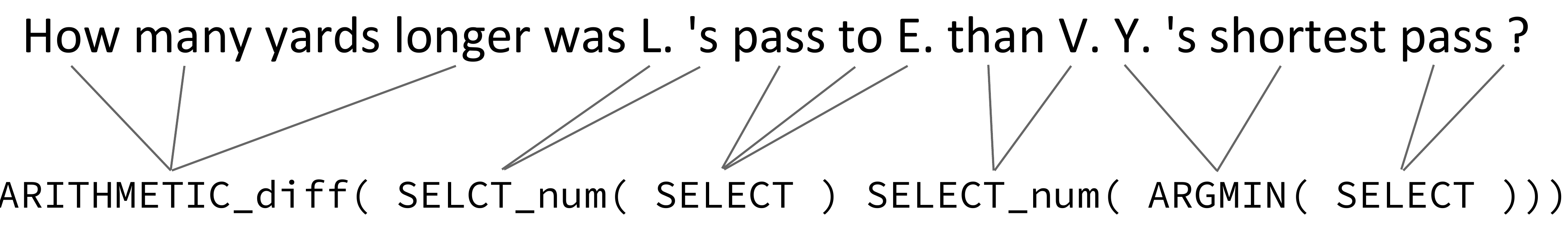}
  \caption{Example of an alignment between question and program tokens in \drop{} as predicted by FastAlign. 
  \nl{Lossman}, \nl{Evan}, \nl{Vince}, and \nl{Young} are abbreviated to \nl{L.}, \nl{E.}, \nl{V.}, and \nl{Y.} for brevity.
  }
  \label{fig:alignment}
\end{figure}

To encourage learning better alignments, we supervise the attention distribution computed by the decoder to attend to specific question tokens at each time-step \cite{liu2016neural}.
We use an off-the-shelf word aligner to produce a ``gold'' alignment between question and program tokens (where program tokens correspond to grammar rules in grammar-based decoding) for all training set examples. Then, at every decoding step where the next prediction symbol's "gold" alignment is to question tokens at indices $\mathcal{I}$, we add the term $-\log \sum_{i \in \mathcal{I}} p_i$ to the objective, pushing the model to put attention probability mass on the aligned tokens. We use the FastAlign word alignment package~\cite{dyer-etal-2013-simple}, based on IBM model 2, which is a generative model that allows to extract word alignments from parallel corpus without any annotated data. Figure~\ref{fig:alignment} shows an example question-program pair and the alignments induced by FastAlign.

\paragraph{(b) Attention over Spans}
Question spans can align to subtrees in the corresponding program. For example, in Fig.~\ref{fig:example_split}, \emph{largest state} aligns to \texttt{state.area = (select max $\ldots$ from state)}. 
Similarly, in a question such as \nl{What does Lionel Messi do for a living?}, the multi-word phrase \nl{do for a living} aligns to the KB relation \texttt{Profession}. Allowing the model to directly attend to multi-token phrases could induce more meaningful alignments that improve compositional generalization.

Here, rather than computing an attention distribution over input tokens $(x_1, \dots x_n)$, we compute a distribution over the set of spans corresponding to all constituents (including all tokens) as predicted by an off-the-shelf constituency parser~\cite{joshi-etal-2018-extending}.
Spans are represented using a self-attention mechanism over the hidden representations of the tokens in the span, as in ~\newcite{lee-etal-2017-coref}.

\paragraph{(c) Coverage}
Questions at test time are sometimes similar to training questions, but include new information expressed by a few tokens. A model that memorizes a mapping from question templates to programs can ignore this new information, hampering compositional generalization. To encourage models to attend to the entire question, we add the attention-coverage mechanism from \newcite{see2017get} to our model.
Specifically, at each decoding step the decoder holds a \emph{coverage vector} $c = (c_1, \dots, c_n)$, where $c_i$ corresponds to the sum of attention probabilities over $x_i$ in all previous time steps. The coverage vector is given as another input to the decoder, and a loss term is added that penalizes attending to tokens with high coverage: $\sum_{i=1}^n \min(c_i, p_i)$, encouraging the model to attend to tokens not yet attended to.

\subsection{Downsampling Frequent Program Templates}
\label{ssec:model-skew}
Training a semantic parser can be hampered if the training data contains a highly skewed distribution over program templates, i.e., a large fraction of the training examples correspond to the same template. 
In such a biased environment, the model might memorize question-to-template mappings instead of modeling the underlying structure of the problem.
We propose to downsample examples from frequent templates such that the resulting training data has a more balanced template distribution. 

Our initial investigation showed that the distribution over program templates in \drop{} is highly skewed ($20$ templates out of $111$ constitute $90\%$ of the data), leading to difficulties to achieve \emph{any} generalization to examples from the program split. Thus, in \drop{}, for any program template in the training set where there are more than $20$ examples, we randomly sample $20$ examples for training.
Downsampling is related to AFLite~\cite{winogrande-sagkuchi-2019, aflite-bras-2020}, an algorithmic approach to bias reduction in datasets. AFLite is applied when bias is hard to define; as we have direct access to a skewed program distribution, we can take a much simpler approach for reducing bias.
\section{Datasets}
\label{sec:datasets}

\begin{table*}[t]
\footnotesize
\begin{tabular}{p{0.96\textwidth}}
\toprule
\textbf{Dataset:} \textbf{\geo{}} \\ \boldmath$x$\textbf{:}\;\;\;\;\textit{how many states border the state with the largest population?} \\
 \boldmath$z$\textbf{:}\;\;\;\;\texttt{select count( border\_info.border ) from border\_info as border\_info where border\_info.state\_name in ( select state.state\_name from state as state where state.population = ( select max( state.population ) from state as state ) )}   \\ 

\midrule
\textbf{Dataset:} \textbf{\atis{}} \\ \boldmath$x$\textbf{:}\;\;\;\;\textit{what is the distance from airport\_code0 airport to city\_name0 ?} \\
\boldmath$z$\textbf{:}\;\;\;\;\texttt{select distinct airport\_service.miles\_distant from airport as airport , airport\_service as airport\_service , city as city where airport.airport\_code = "airport\_code0" and airport.airport\_code = airport\_service.airport\_code and city.city\_code = airport\_service.city\_code and city.city\_name = "city\_name0"}   \\ 

\midrule
\textbf{Dataset:} \textbf{\scholar{}} \\ \boldmath$x$\textbf{:}\;\;\;\;\textit{What papers has authorname0 written?} \\
\boldmath$z$\textbf{:}\;\;\;\;\texttt{select distinct paper.paperid from author as author , paper as paper , writes as writes where author.authorname = "authorname0" and writes.authorid = author.authorid and writes.paperid = paper.paperid}   \\ 
\midrule
\textbf{Dataset:} \textbf{\advising{}} \\ \boldmath$x$\textbf{:}\;\;\;\;\textit{Can undergrads enroll in the course number0 ?} \\
\boldmath$z$\textbf{:}\;\;\;\;\texttt{select distinct course.advisory\_requirement , course.enforced\_requirement , course.name from course as course where course.department = "department0" and course.number = number0}   \\
\midrule
\textbf{Dataset:} \textbf{\drop{}} \\ \boldmath$x$\textbf{:}\;\;\;\;\textit{How many yards longer was Johnson's longest touchdown compared to his shortest touchdown of the first quarter?} \\
\boldmath$z$\textbf{:}\;\;\;\;\texttt{ARITHMETIC\_diff( SELECT\_num( ARGMAX( SELECT ) ) SELECT\_num( ARGMIN( FILTER( SELECT ) ) ) )}   \\
\bottomrule
\end{tabular}
\caption{Examples for the different datasets, of a question ($x$) and its corresponding program ($z$).}
\label{tab:dataset_examples}
\end{table*}

We create iid and program splits for five datasets according to the procedure of \newcite{finegan-dollak-etal-2018-improving} as described in \S\ref{sec:comp_gen}:\footnote{We do not use their original split because we remove duplicate question-program pairs and balance the number of examples between the iid and program splits.} Four text-to-SQL datasets from \newcite{finegan-dollak-etal-2018-improving} and one dataset for mapping questions to QDMR programs \cite{break2020} in DROP~\cite{drop2019}. Similar to prior work \cite{finegan-dollak-etal-2018-improving}, we train and test models on anonymized programs, that is, entities are replaced with typed variables (\S\ref{sec:comp_gen}).
Table~\ref{tab:dataset_examples} gives an example question and program for each of these datasets. 
\begin{itemize}[leftmargin=*,topsep=0pt,itemsep=0pt,parsep=0pt]
    \item \textbf{\atis}: questions for a flight-booking task \cite{price1990evaluation, dahl1994expanding}.
    \item \textbf{\geo}: questions about US geography \cite{zelle1996learning}.
    \item \textbf{\advising}: questions about academic course information. \cite{finegan-dollak-etal-2018-improving}. 
    \item \textbf{\scholar}: questions about academic publications \cite{iyer2017neural}.
    \item \textbf{\drop}: questions on history and football games described in text paragraphs. We use annotated QDMR programs from~\citet{break2020}.
\end{itemize}

\paragraph{SQL Grammar:} We adapt the SQL grammar developed for \atis{} \cite{lin2019grammar} to cover the four SQL datasets. To achieve that, additional data normalization steps were taken (see appendix), such as rewriting programs to have a consistent SQL style. The grammar uses the DB schema to produce domain-specific production rules, e.g., in \atis{} \texttt{table\_name $\rightarrow$ FLIGHTSalias0}, \texttt{column\_name $\rightarrow$ FLIGHTSalias0.MEAL\_DESCRIPTION}, and \texttt{value $\rightarrow$ class\_type0}. At inference time, we enforce context-sensitive constraints that eliminate production rules that are invalid given the previous context. For example, in the \texttt{WHERE} clause, the set of \texttt{column\_name} rules is limited to columns that are part of previously mentioned tables. These constraints reduce the number of syntactically invalid programs, but do not eliminate them completely.

\paragraph{DROP Grammar:} 
We manually develop a grammar over QDMR programs to perform grammar-based decoding for \drop{}, similar to ~\citet{gupta-etal-2020-nmn}.
This grammar contains typed operations required for answering questions, such as, \texttt{ARITHMETIC\_diff(NUM, NUM) $\rightarrow$  NUM}, \texttt{SELECT\_num(PassageSpan) $\rightarrow$ NUM}, and \texttt{SELECT $\rightarrow$ PassageSpan}.
Because QDMR programs are executed over text paragraphs (rather than a KB), 
QDMR operators receive string arguments as inputs (analogous to KB constants), which we remove for anonymization (Table~\ref{tab:dataset_examples}). This results in program templates that include only the logical operations required for finding the answer.
While such programs cannot be executed as-is on a database, they are sufficient for the purpose of testing compositional generalization in semantic parsing, and can be used as ``layouts'' in a neural module network approach \cite{gupta-etal-2020-nmn}.

\section{Experiments}
\label{sec:experiments}

We now present our empirical evaluation of compositional generalization.

\subsection{Experimental Setup}

\begin{table}[t]
\centering
\resizebox{1.0\columnwidth}{!}{
\begin{tabular}{lccc}
\toprule
\multirow{2}{*}{\textbf{Dataset}} & \multirow{2}{*}{\textbf{Split}}   & \textbf{\# examples}           & \textbf{\# new templates} \\ 
                         &                          & (train / dev / test)  & (train / dev / test) \\
\midrule
\multirow{2}{*}{\textsc{GeoQuery}} & iid  & 409  / 103 / 95      & 192 / 32  / 24  \\
                                   & Prog.  & 424  / 91  / 91      & 148 / 49  / 47  \\ \midrule
\multirow{2}{*}{\textsc{Atis}}     & iid  & 3014 / 405 / 402     & 830 / 48 / 65 \\
                                   & Prog.  & 3061 / 373 / 375     & 645 / 140 / 148 \\ \midrule
\multirow{2}{*}{\textsc{Scholar}}  & iid  & 433  / 111 / 105     & 158 / 16  / 16  \\
                                   & Prog.  & 454  / 97  / 98      & 112 / 37  / 37  \\ \midrule
\multirow{2}{*}{\textsc{Advising}} & iid  & 3440 / 451 / 446     & 203 / 0 / 0 \\
                                   & Prog.  & 3492 / 421 / 414     & 163 / 20  / 17  \\ \midrule
\multirow{2}{*}{\textsc{DROP}}     & iid  & 582  / 102 / 500     & 73  / 0  / 0  \\                                                                 
                                   & Prog.  & 582  / 102 / 385     & 73  / 0  / 38  \\ 
\bottomrule
\end{tabular}}
\caption{\label{tab:stats} Dataset statistics for the iid split and the program (prog.) split for all datasets. \textit{\# new templates} indicates the number of templates unseen during training time for the development and test sets, and the total number of templates for the training set.}
\end{table}

\label{sec:data_processing}
We create training/development/test splits using both an \emph{iid split} and a \emph{program split}, such that the number of examples is similar across splits.
Table~\ref{tab:stats} presents exact statistics on the number of unique examples and program templates for all datasets. There are much fewer new templates in the development and test sets for the iid split than for the program split, thus the iid split requires less compositional generalization. In \drop{}, we report results for the downsampled dataset (\S\ref{ssec:model-skew}), and analyze downsampling below.

\paragraph{Evaluation Metric} We evaluate models using exact match (EM), that is, whether the predicted program is identical to the gold program. In addition, we report \emph{relative gap}, defined as $1 - \frac{\emprog}{\emiid}$, where $\emprog$ and $\emiid$ are the EM on the program and iid splits, respectively. This metric measures the gap between in-distribution generalization and OOD generalization, and our goal is to minimize it (while additionally maximizing $\emiid$).

We select hyper-parameters by tuning the learning rate, batch size, dropout, hidden dimension, and use early-stopping w.r.t. development set EM (specific values are in the appendix). The results reported are averaged over 5 different random seeds.

\paragraph{Evaluated Models}
Our goal is to measure the impact of various modeling choices on compositional generalization. 
We term our baseline sequence-to-sequence semantic parser \seqtoseq{}, and denote the parser that uses grammar-based decoding by \grammar{} (\S\ref{ssec:model-grammar}).
Use of contextualized representations in these parsers is denoted by +\elmo{} and +\bert{} (\S\ref{ssec:model-repr}). 
We also experiment with the proposed additions to the decoder attention~(\S\ref{ssec:model-attn}). In a parser, use of (a) auxiliary attention supervision obtained from FastAlign is denoted by +\attnsup{}, (b) use of attention over constituent spans by +\attnspan{}, and (c) use of attention-coverage mechanism by +\coverage{}.

\begin{table}[tb]
\centering
\footnotesize
\begin{tabular}{lccc}
\toprule
\multirow{2}{*}{\textbf{Model}}         & \textbf{iid} & \textbf{Program} & \textbf{Rel.}\\
                                        & \textbf{split}  & \textbf{split} & \textbf{gap} \\
\midrule
SQL                            &            &           &      \\          
          \;\;\seqtoseq{}      & 74.9       & 10.8      & 84.6 \\
          \;\;\;\;+\elmo{}     & 76.2      & \textbf{15.9}      & \textbf{77.9} \\
          \;\;\;\;+\bert{}     & \textbf{77.5}       & 10.5      & 85.7 \\
          \addlinespace[2pt] 
          \;\;\grammar{}       & 70.1       & 14.1       & 78.1 \\
          \;\;\;\;+\elmo{}     & 65.5       & 11.2      & 81.4 \\
          \;\;\;\;+\bert{}     & 67.6       & 8.4      & 86.7 \\

\midrule
DROP                           &            &           &      \\          
          \;\;\seqtoseq{}      & 45.4       & 1.6       & 96.4 \\
          \;\;\;\;+\elmo{}     & 53.2       & 2.1       & 96.0 \\
          \;\;\;\;+\bert{}     & 50.0       & 0.0       & 100  \\
          \addlinespace[2pt]
          \;\;\grammar{}       & 49.2       & 2.6       & 94.7 \\
          \;\;\;\;+\elmo{}     & 57.8       & \textbf{13.2}       & \textbf{77.1} \\
          \;\;\;\;+\bert{}     & \textbf{64.6}       & 3.9       & 93.9 \\
\bottomrule
\end{tabular}
\caption{\label{tab:conrepr} Test results for contextualized representations and grammar-based decoding.
}
\end{table}

\begin{table}[tb]
\centering
\footnotesize
\begin{tabular}{lccc}
\toprule
\multirow{2}{*}{\textbf{Model}}         & \textbf{iid} & \textbf{Program} & \textbf{Rel.}\\
                                        & \textbf{split}  & \textbf{split} & \textbf{gap} \\
\midrule
SQL                                     &           &           &      \\          
          \;\;\seqtoseq{}               & 74.9      & 10.8      & 84.6 \\
          \;\;\;\;+\attnsup{}           & 73.3      & 18.5      & 73.2 \\
          \;\;\;\;+\elmo{}              & \textbf{76.2}      & 15.9           & 77.9 \\
          \;\;\;\;+\elmo{}+\attnsup{}   & 73.7      & \textbf{20.3}           & \textbf{70.6} \\
          \addlinespace[3pt]
          \;\;\grammar{}                & 70.1       & 14.1       & 78.1 \\
          \;\;\;\;+\attnsup{}           & 73.3      & 15.8      & 75.3 \\
          \;\;\;\;+\elmo{}              & 65.5       & 11.2      & 81.4 \\
          \;\;\;\;+\elmo+\attnsup{}     & 69.1      & 11.8      & 81.6 \\
\midrule
DROP                                    &           &           & \\          
          \;\;\seqtoseq{}               & 45.4      & 1.6       & 96.4 \\
          \;\;\;\;+\attnsup{}           & 49.4      & 1.3       & 97.3 \\
          \;\;\;\;+\elmo{}              & 53.2      & 2.1       & 96.0 \\
          \;\;\;\;+\elmo{}+\attnsup{}   & 58.2      & 2.6       & 95.5 \\
          \addlinespace[3pt]
          \;\;\grammar{}                & 49.2      & 2.6       & 94.7 \\
          \;\;\;\;+\attnsup{}           & 55.8      & 4.7       & 91.5 \\
          \;\;\;\;+\elmo{}              & 57.8      & \textbf{13.2}       & \textbf{77.1} \\
          \;\;\;\;+\elmo{}+\attnsup{}   & \textbf{59.8}      & 12.2      & 79.5 \\
\bottomrule
\end{tabular}
\caption{\label{tab:attnsup} Test results for auxiliary attention supervision.}
\end{table}

\begin{table}[tb]
\centering
\footnotesize
\begin{tabular}{lccc}
\toprule
\multirow{2}{*}{\textbf{Model}}         & \textbf{iid} & \textbf{Program} & \textbf{Rel.}\\
                                        & \textbf{split}  & \textbf{split} & \textbf{gap} \\
\midrule
SQL                                     &           &           &      \\          
          \;\;\seqtoseq{}               & 74.9      & 10.8      & 84.6 \\
          \;\;\;\;+\coverage{}           & 75.3      & 17      & 76.2 \\
          \;\;\;\;\;\;+\attnsup{} & 72.4      &  23.5    & 65.8 \\
          \;\;\;\;+\elmo{}              & 76.2      & 15.9           & 77.9 \\
          \;\;\;\;+\elmo{}+\coverage{}   & \textbf{76.2}      & 24.1          & 66.5 \\
          \;\;\;\;\;\;+\attnsup{} & 72      &  \textbf{25.4}    & \textbf{62.2} \\
\midrule
DROP                                    &           &           & \\          
          \;\;\seqtoseq{}               & 45.4      & 1.6       & 96.4 \\
          \;\;\;\;+\coverage{}          & 47.2      &  2.1      & 95.5 \\
          \;\;\;\;+\elmo{}              & 53.2      & 2.1       & 96.0 \\
           \;\;\;\;+\elmo{}+\coverage{}   & \textbf{64.4}   & \textbf{4.4}     & \textbf{93.1} \\
\bottomrule
\end{tabular}
\caption{\label{tab:attncoverage} Test results for attention-coverage.}
\end{table}

\subsection{Main Results}
Below we present the performance of our various models on the test set, and discuss the impact of these modeling choices.
For SQL, we present results averaged across the four datasets, and report the exact numbers for each dataset in Table~\ref{tab:main_verbose_results}.

\paragraph{Baseline Performance} The top-row in Table~\ref{tab:conrepr} shows the performance of our baseline \seqtoseq{} model using GloVe representations. In SQL, it achieves $74.9$ EM on the iid split and $10.8$ EM on the program split, and in DROP, $45.4$ EM and a surprisingly low $1.6$ EM on the iid and program splits, respectively. A possible reason for the low program split performance on \drop{} is that programs include only logical operations without any KB constants (\S\ref{sec:datasets}), making generalization to new compositions harder than in SQL (see also analysis in \S\ref{subsec:analysis}).
As observed by \citet{finegan-dollak-etal-2018-improving}, there is a large relative gap in performance on the iid vs. program split.

\paragraph{Contextualized Representations} Table~\ref{tab:conrepr} shows that contextualized representations consistently improve absolute performance and reduce the relative gap in \drop{}. In SQL, contextualized representations improve absolute performance and reduce the relative gap in the \seqtoseq{} model, but not in the \grammar{} model. The relative gap is reduced by roughly $7$ points in SQL, and $17$ points in \drop{}. As \elmo{} performs slightly better than \bert{},
we present results only for \elmo{} in some of the subsequent experiments, and report results for \bert{} in Table~\ref{tab:main_verbose_results}.

\paragraph{Grammar-based Decoding} 
Table~\ref{tab:conrepr} shows that grammar-based decoding both increases accuracy and reduces the relative gap on \drop{} in all cases. In SQL, grammar-based decoding consistently decreases the absolute performance compared to \seqtoseq{}. We conjecture this is because our SQL grammar contains a large set of rules meant to support the normalized SQL structure of \newcite{finegan-dollak-etal-2018-improving}, which makes decoding this structure challenging. We provide further in-depth comparison of performance in \S\ref{subsec:analysis}.

\begin{table}[tb]
\centering
\footnotesize
\begin{tabular}{lccc}
\toprule
\multirow{2}{*}{\textbf{Model}}         & \textbf{iid}    & \textbf{Program} & \textbf{Rel.}\\
                                        & \textbf{split}  & \textbf{split}   & \textbf{gap} \\
\midrule
SQL                                     &                 &                  &                   \\          
          \;\;\seqtoseq{}               & 74.9      & 10.8      & 84.6 \\
          \;\;\;\;+\attnspan{}        &  73.8       &  14.3    & 79.5 \\
          \;\;\;\;+\elmo{}              & \textbf{76.2}      & 15.9           & 77.9 \\
          \;\;\;\;+\elmo{}+\attnspan{} &  75.5       &  \textbf{16.3}    & \textbf{77.2} \\
          
\midrule
DROP                                    &                 &                  &                   \\          
          \;\;\seqtoseq{}               & 45.4      & 1.6     & 96.4 \\
          \;\;\;\;+\attnspan{}          & 48.6     & \textbf{3.1}   & \textbf{93.6} \\
          \;\;\;\;+\elmo{}              & 53.2      & 2.1     & 96.0 \\
          \;\;\;\;+\elmo{} +\attnspan{}  & \textbf{56.2}     & 1.6     & 97.1 \\
\bottomrule
\end{tabular}
\caption{\label{tab:attnspan} Test results for attention over spans.}
\end{table}

\begin{table}[tb]
\centering
\footnotesize
\begin{tabular}{lcccc}
\toprule
\multirow{2}{*}{\textbf{Model}}     & \multicolumn{2}{c}{\textbf{iid split }} &\multicolumn{2}{c}{\textbf{Program split }}\\
                                    \cmidrule(lr){2-3} \cmidrule(lr){4-5} 
                                    & w/o DS    & w/ DS         & w/o DS    & w/ DS            \\
\midrule
          \seqtoseq{}                 & \textbf{49.8}  & 45.4           & 0.0      & \textbf{1.6}        \\
          \grammar{}                  & \textbf{51.6}  & 49.2           & 0.0      & \textbf{2.6}        \\
          \;\;+\elmo{}                & 52.8           & \textbf{57.8}  & 0.8      & \textbf{13.2}        \\
\bottomrule
\end{tabular}
\caption{\label{tab:downsampling} Reducing training data bias in DROP by downsampling examples for frequent templates leads to better compositional generalization in all models. }
\end{table}

\paragraph{Attention Supervision}
Table~\ref{tab:attnsup} shows that attention supervision has a substantial positive effect on compositional generalization, especially in SQL. In SQL, adding auxiliary attention supervision to a \seqtoseq{} model improves the program split EM from $10.8~\rightarrow~18.5$, and combining with \elmo{} leads to an EM of $20.3$. Overall, using \elmo{} and \attnsup{} reduces the relative gap from $84.6 \rightarrow 70.6$ compared to \seqtoseq{}.
In \drop{}, attention supervision improves iid performance and reduces the relative gap for \grammar{} using GloVe representations, but does not lead to additional improvements when combined with \elmo{}. 

\paragraph{Attention-coverage}
Table~\ref{tab:attncoverage} shows that attention-coverage improves absolute performance and compositional generalization in all cases. Interestingly, in SQL, best results are obtained without the attention coverage loss term, but still providing the coverage vector as additional input to the decoder. 
In SQL, adding attention-coverage improves program split EM from $10.8 \rightarrow 17$. Combining coverage with \elmo{} and \attnsup{} leads to our best results, where program split EM reaches $25.4$, and the relative gap drops from $84.6 \rightarrow 62.2$ (with a slight drop in iid split EM).
In \drop{}, using attention-coverage mechanism with auxiliary coverage loss improves iid performance from $53.2~\rightarrow~64.6$ and reduces the relative gap from $96~\rightarrow~93.1$.

\paragraph{Attention over Spans}
Table~\ref{tab:attnspan} shows that, without \elmo{}, attention over spans improves iid and program split EM in both SQL and \drop{}, but when combined with \elmo{} differences are small and inconsistent.

\paragraph{Downsampling Frequent Templates} Table~\ref{tab:downsampling} shows that for DROP, where the distribution over program templates is extremely skewed, downsampling training examples for frequent templates leads to better compositional generalization in all models. For example, without downsampling (w/o DS), program split EM drops from $13.2 \rightarrow 0.8$ for the \grammar{}+\elmo{} model. 

\begin{table}[t]
\centering
\footnotesize
\resizebox{1.0\columnwidth}{!}{
\begin{tabular}{l|c|c|c}
\toprule
\multirow{2}{*}{\textbf{Model}}          & \textbf{Seen} & \textbf{New} & \textbf{Invalid}\\
                                         & \textbf{program} & \textbf{program} & \textbf{syntax}\\
\midrule
          \seqtoseq{}              & 75.7 & 19.6 & \textbf{4.7} \\ 
           \;\;+\elmo{}            & 64.9 & 26.2 & 8.9 \\ 
          \;\;+\attnsup{}       & 62.6 & 29 & 8.3 \\ 
          \;\;\;\;+\elmo        & 57.4 & 32.4 & 10.2 \\ 
          \;\;+\coverage{}       & 59.8 & 28.9 & 11.3 \\ 
          \;\;\;\;+\elmo         & \textbf{40.5} & \textbf{41.3} & 18.1 \\ 
        \;\;+\attnspan{}        & 70.2 & 22.2 & 7.5 \\ 
          \;\;\;\;+\elmo          & 63.1 & 29.3 & 7.6 \\ 
          \addlinespace[2pt]
          \midrule
          \grammar{}         & 26.2 & 70.4 & \textbf{3.4} \\ 
          \;\;+\elmo{}        & \textbf{22} & \textbf{71.7} & 6.3 \\ 
          \;\;+\attnsup{}        & 25.7 & 68.6 & 5.7 \\ 
          \;\;\;\;+\elmo         & 26.8 & 69.3 & 3.9 \\ 
\bottomrule
\end{tabular}}
\caption{\label{tab:sqlerrors} 
Analysis of program split development set results across all SQL datasets. 
}
\end{table}

\renewcommand{\arraystretch}{1.0}
\begin{table*}[t]
\centering
\footnotesize
\resizebox{\textwidth}{!}{
\begin{tabular}{l|ccc|ccc|ccc|ccc}
\toprule
\multirow{2}{*}{\textbf{Model}}         & \multicolumn{3}{c}{Advising} & \multicolumn{3}{c}{ATIS} & \multicolumn{3}{c}{GeoQuery} & \multicolumn{3}{c}{Scholar}\\
                                        &  iid split & Prog. split & Rel. gap &  iid split & Prog. split & Rel. gap &  iid split & Prog. split & Rel. gap &  iid split & Prog. split & Rel. gap\\

\midrule
\seqtoseq{}           &                90.0 &                0.1 &     99.9 &                  70.5 &                 12.3 &       82.6 &                  70.1 &                 19.1 &       72.8 &                  69.1 &                 11.6 &       83.2 \\
\;\;+\elmo              &                91.7 &                1.9 &     97.9 &                  71.6 &                 21.1 &       70.5 &                  73.7 &                 27.9 &       62.1 &                  68.0 &                 12.9 &       81.0 \\
\;\;+\bert              &                91.5 &                0.1 &     99.9 &                  72.2 &                 17.0 &       76.5 &                  74.7 &                 18.0 &       75.9 &                  71.4 &                  6.7 &       90.6 \\
\midrule
\;\;+\attnsup{}         &                87.4 &                1.1 &     98.7 &                  69.6 &                 28.3 &       59.3 &                  72.4 &                 25.9 &       64.2 &                  64.0 &                 18.8 &       70.6 \\
\;\;\;\;+\elmo{}            &                89.1 &                0.4 &     99.6 &                  71.4 &                 28.3 &       60.4 &                  71.6 &                 \textbf{32.5} &       54.6 &                  62.7 &                20.2 &       67.8 \\
\;\;\;\;+\bert{} &                90.2 &                2.3 &     97.5 &                  70.1 &                 29.9 &       57.3 &                  74.7 &                 29.2 &       60.9 &                  64.8 &                 16.3 &       74.8 \\
\;\;+\coverage{}    &                90.1 &                1.9 &     97.9 &                  70.7 &                 23.7 &       66.5 &                  72.4 &                 27.7 &       61.7 &                  67.8 &                 14.5 &       78.6 \\
\;\;\;\;+\elmo{} &                91.9 &                5.4 &     94.1 &                  \textbf{74.5} &                 \textbf{34.4} &       53.8 &                  73.3 &                 28.4 &       61.3 &                  65.1 &                 \textbf{28.2} &       56.7 \\
\;\;\;\;+\bert{} &                \textbf{92.4} &                5.0 &     94.6 &                  73.6 &                 31.7 &       56.9 &                  \textbf{76.6} &                 28.6 &       62.7 &                  \textbf{73.0} &                 23.9 &       67.3 \\
\;\;\;\;+\attnsup{}  & 85.9 & 3.2  & 96.3 & 71.1 & 31.4 & 55.8 & 72.6 & \textbf{34.7} & 52.2 & 60 & 24.7 & 58.8  \\  
\;\;\;\;\;\;+\elmo{}  & 88.6 & 5  & 94.4 & 71 & 34.3 & \textbf{51.7} & 70.5 & 34.1 & \textbf{51.6} & 57.7 & \textbf{28.2} & \textbf{51.1}  \\  
\;\;\;\;\;\;+\bert{}  & 89.1 & 4.9 & 94.5 & 71.8 & 33.6 & 53.2 & 73.9 & 31.6 & 57.2 & 63.2 & 27.6 & 56.3   \\  
\;\;+\attnspan{}        &                89.3 &                3.4 &     96.2 &                  70.4 &                 17.9 &       74.6 &                  70.5 &                 22.2 &       68.5 &                  65.1 &                 13.9 &       78.6 \\
\;\;\;\;+\elmo{}            &                92.2 &                4.8 &     94.8 &                  72.4 &                 23.5 &       67.5 &                  69.5 &                 24.8 &       64.3 &                  67.8 &                 12.2 &       82.0 \\
\;\;\;\;+\bert{} &                91.9 &                0.0 &    100.0 &                  71.5 &                 22.6 &       68.4 &                  72.0 &                 21.1 &       70.7 &                  65.3 &                  9.4 &       85.6 \\
\midrule
\grammar{}            &                88.5 &                3.0 &     96.6 &                  65.8 &                 18.1 &       72.5 &                  63.2 &                 21.8 &       65.5 &                  61.1 &                 13.7 &       77.6 \\
\;\;+\elmo              &                90.0 &                3.1 &     96.6 &                  61.3 &                 21.3 &       65.3 &                  58.1 &                 16.3 &       71.9 &                  52.6 &                  4.3 &       91.8 \\
\;\;+\bert              &                90.7 &                2.3 &     97.5 &                  62.4 &                  7.1 &       88.6 &                  63.2 &                 20.0 &       68.4 &                  54.1 &                  4.1 &       92.4 \\
\midrule
\;\;+\attnsup{}        &                87.4 &                \textbf{5.9} &     \textbf{93.2} &                  63.8 &                 24.2 &       62.1 &                  64.2 &                 20.4 &       68.2 &                  63.8 &                 14.3 &       77.6 \\
\;\;\;\;+\elmo            &                89.1 &                2.0 &     97.8 &                  65.0 &                 15.9 &       75.5 &                  62.9 &                 22.4 &       64.4 &                  59.2 &                  6.7 &       88.7 \\
\;\;\;\;+\bert &                89.8 &                3.5 &     96.1 &                  61.4 &                  3.5 &       94.3 &                  66.5 &                 12.5 &       81.2 &                  54.3 &                  3.9 &       92.8 \\
\bottomrule
\end{tabular}
}
\caption{\label{tab:main_verbose_results} Test EM for all models and SQL datasets. All results are averages over $5$ different random seeds.}
\end{table*}

\paragraph{Takeaways}
We find that contextualized representations, attention supervision, and attention coverage generally improve program split EM and reduce the relative gap, perhaps at a small cost to iid split EM. In \drop{}, grammar-based decoding is important, as well as downsampling of frequent templates. 
Overall the gap between in-distribution and OOD performance dropped from $84.6 \rightarrow 62.2$ for SQL, and from $96.4 \rightarrow 77.1$ for \drop{}. While this improvement is significant, it leaves much to be desired in terms of models and training procedures that truly close this gap.

\subsection{Analysis}
\label{subsec:analysis}
\paragraph{Error Analysis}
We analyze the errors of each model on the program split development set for all SQL datasets and label each example with one of three categories
(Table~\ref{tab:sqlerrors}): 
\textit{Seen programs} are errors resulting from outputting program templates that appear in the training set,
while \textit{new programs} are wrong programs that were not observed in the training set. \textit{Invalid syntax} errors are outputs that are syntactically invalid programs.
Table~\ref{tab:sqlerrors} shows that for \seqtoseq{} models, those that improve compositional generalization also increase the frequency of \textit{new programs} and \textit{invalid syntax} errors. 
Grammar-based models output significantly more \textit{new programs} than \seqtoseq{} models, and less \textit{invalid syntax} errors.\footnote{The grammar can still produce invalid outputs (see~\S\ref{sec:datasets} - SQL Grammar), thus it does not eliminate these errors entirely.} Overall, the correlation between successful compositional generalization and the rate of \textit{new programs} is inconsistent.

\begin{table}[t]
\centering
\footnotesize
\resizebox{1.0\columnwidth}{!}{
\begin{tabular}{l|c|c|c|c}
\toprule
\multirow{2}{*}{\textbf{Model}}          & \textbf{Semantically} & \textbf{Semantically} & \textbf{Limited} & \textbf{Significant}\\
                                         & \textbf{equivalent} & \textbf{similar} & \textbf{divergence} & \textbf{divergence}\\
\midrule
\textbf{program split} & & & & \\
\addlinespace[2pt]
        \;\;\seqtoseq{}+\elmo{}    & 4 & 7 & 4 & 15 \\
        \;\;+\attnsup{}            & 7 & 7 & 5 & 11 \\
        \;\;+\coverage{}           & 4 & 11 & 2 & 13 \\
        \;\;+\attnspan{}           &  5 & 9  &  0 & 16 \\
\midrule
\textbf{iid split} & & & &\\
\addlinespace[2pt]
          \;\;\seqtoseq{}             & 6 & 8 & 4  & 12 \\ 
          \;\;\grammar{}              & 6 & 11 & 7 & 6\\ 
\bottomrule
\end{tabular}}
\caption{\label{tab:sqlerrorsmanual} Manual categorization of 30 random predictions on the iid and program splits development sets.}
\end{table}

We further inspect 30 random predictions of multiple models on both the program split and the iid split (Table~\ref{tab:sqlerrorsmanual}). \textit{Semantically equivalent} errors are predictions that are equivalent to the target programs. \textit{Semantically similar} is a relaxation of the former category (e.g., an output that represents \nl{flights that depart at time0}, where the gold program represents \nl{flights that depart after time0}). \textit{Limited divergence} or \textit{significant divergence} corresponds to invalid programs that slightly or significantly  diverge from the  target output, respectively.

Table~\ref{tab:sqlerrorsmanual} shows that adding attention-supervision, attention-coverage, and attention over spans increases the number of predictions that are semantically close to the target programs.
We also find that the frequency of correct typed variables in predictions is significantly higher when using attention-supervision and attention-coverage compared to the baseline model ($p < 0.05$). 
In addition, the errors of the \grammar{} model tend to be closer to the target program compared to \seqtoseq{}.

\paragraph{Compositional Generalization in \drop{}}
Our results show that compositional generalization in \drop{} is harder than in the SQL datasets. We hypothesized that this could be due to the existence of KB relations in SQL programs after program anonymization, while QDMR programs do not contain any arguments. To assess that, we further anonymize the predicates in all SQL programs in all four datasets, such that the SQL programs do not contain any KB constants at all (similar to \drop{}). We split the data based on this anonymization, and term it the \emph{KB-free split}.
On the development set, when moving from a program split to a KB-free split, the average accuracy drops from $14.5 \rightarrow 9.8$. This demonstrates that indeed a KB-free split is harder than the program split from \newcite{finegan-dollak-etal-2018-improving}, partially explaining the difference between SQL and \drop{}.

\section{Conclusion}
We presented a comprehensive evaluation of \emph{compositional generalization} in semantic parsers by analyzing the performance of a wide variety of models across 5 different datasets.
We experimented with well-known extensions to sequence-to-sequence models and also proposed novel extensions to the decoder's attention mechanism.
Moreover, we proposed reducing dataset bias towards a heavily skewed program template distribution by downsampling examples from frequent templates.

We find that our proposed techniques improve generalization to OOD examples.
However, the generalization gap between in-distribution and OOD data remains high. 
This suggests that future research in semantic parsing should consider more drastic changes to the prevailing encoder-decoder approach to address compositional generalization.

\section*{Acknowledgements}
This research was supported by The Israel Science Foundation (grant 942/16), 
The Yandex Initiative for Machine Learning, 
The European Research Council (ERC) under the European Union Horizons 2020 research and innovation programme (grant ERC DELPHI 802800), 
and the Army Research Office (grant number W911NF-20-1-0080).
The second author was supported by a Google PhD fellowship.

\bibliography{references}
\bibliographystyle{acl_natbib}

\newpage
\appendix
\section{SQL Style}
SQL programs vary in style across datasets. We address a specific difference concerning the syntax to neutralize an interaction with the models analyzed in this analysis, and allow comparability across models and datasets. We standardize the form {\fontfamily{qcr}\selectfont
<table1> <join> <table2> ON <condition>}
 by replacing {\fontfamily{qcr}\selectfont<join>} with a comma and adding {\fontfamily{qcr}\selectfont<condition>} to the {\fontfamily{qcr}\selectfont WHERE} clause.

\section{SQL Grammar Development}
Our SQL grammar is a context-free grammar. We fit an existing implementation for text-to-SQL ~\cite{lin2019grammar} to the datasets we experimented with. Examples for grammar rules are in ~Table~\ref{tab:grammars}. At each step, a sequence of non-terminal or terminal expressions (right side) is derived from some non-terminal (left side). 

The SQL programs in the text-to-SQL datasets have aliases for all tables, sub-queries, and custom fields. Also, each column in the program is preceded by an aliased table or a sub-query. To allow the model to generate all aliases, we add terminal rules based on the dataset schema. We modify the rules to create sub-queries and fields so that the use of aliases is enforced, and we add the alias patterns for custom field and tables. We add the table names in the schema, concatenated with the alias patterns, to \textit{table\_name}. We define \textit{col\_ref} as the concatenation of an aliased table and a column of this table. Additionally, we add valid combinations of aliased variables and schema entities.

To allow comparability with \seqtoseq{} models, we use only examples that are parsed by the grammar in the development and test sets, eliminating $39$ examples from \advising{}, $18$ from \atis{} and one example from \geo{}. The grammar covers at least $95\%$ of each train set.

During inference we enforce contextual rules. For example, forcing the derivation of \textit{from\_clause} to have the tables that were selected in \textit{select\_results}. We check validity by executing the programs against the dataset database in Mysql server $5.7$. Some of the programs in our datasets were not executable due to inconsistent use of aliases, or partial column references. We were not able to automatically fix all the programs. We relaxed our constraints to allow the generation of all target programs, hence allowing some invalid outputs.

\begin{table}[t]
\footnotesize
\resizebox{1.0\columnwidth}{!}{
\begin{tabular}{l|c|c|c|c}
\toprule
\textbf{Model} & \advising{} & \atis{} & \geo{} & \scholar{} \\
\midrule
\seqtoseq{}             & 0.7 & 2.4 & 0.4 & 0.2 \\      
\;\;+\elmo              & 0.8 & 7.1 & 0.3 & 0.6 \\           
\;\;+\bert              & 1.7 & 3.3 & 0.3 & 0.5 \\          
\midrule
\;\;+\attnsup{}         & 3.2 & 6.0 &  0.6  & 0.6 \\      
\;\;\;\;+\elmo          & 0.6   & 4.2   & 0.6 & 1.2 \\          
\;\;+\coverage          & 8.0   & 11.9  & 0.6 & 1.0 \\
\;\;+\attnspan{}        & 4.0   & 6.1   & 0.5   & 0.6 \\        
\;\;\;\;+\elmo          & 4.1   & 7.2   & 0.7   & 0.7 \\       
\midrule
\grammar{}              & 18.8 & 25.5 &  1.7 & 0.6 \\         
\;\;+\elmo              & 8.8 & 22.1 &  0.8 & 1.0 \\        
\;\;+\bert              & 20.7 & 36.0 &  1.5 & 1.4 \\         
\midrule
\;\;+\attnsup{}         & 25.6 & 37.3 & 1.8 & 1.5 \\       
\;\;\;\;+\elmo          & 28.6 & 40.3 & 6.2 & 2.4 \\          
\bottomrule
\end{tabular}}
\caption{Average training duration in hours for models trained on SQL datasets.}
\label{tab:runtime}
\end{table}

\begin{table*}[ht]
\footnotesize
\begin{tabular}{p{0.96\textwidth}}
\toprule
\textbf{Global structure}  \\
\;\;\textit{query}\;\;\;\;\;\;\;\;\;\; \textit{select\_core}, \textit{groupby\_clause}, \textit{orderby\_clause}, \;\;\textit{limit} \\
\;\;\textit{select\_core}\;\;\;\;\;\; \textit{select\_with\_distinct }, \textit{select\_results}, \;\;\textit{from\_clause}, \texttt{"WHERE"}, \textit{where\_clause} \\
\midrule
\textbf{Select clause}  \\
\;\;\textit{select\_results}\;\;\;\;\; \textit{select\_result}, \texttt{","}, \textit{select\_result} \\
\;\;\textit{select\_result}\;\;\;\;\;\; \textit{function} \\
\midrule
\textbf{From clause}  \\
\;\;\textit{source}\;\;\;\;\;\;\;\;\;\textit{single\_source}, \texttt{","}, \textit{source} \\
\;\;\textit{single\_source}\;\;\;\;\;\textit{source\_subq} \\
\;\;\textit{source\_subq}\;\;\;\;\;\;\texttt{"("}, \textit{query},\texttt{")"}, \texttt{"AS"}, \textit{subq\_alias} \\
\;\;\textit{source\_table}\;\;\;\;\;\texttt{"TABLE\_PLACEHOLDER"}, \texttt{"AS"}, \textit{table\_name} \\
\midrule
\textbf{Where clause}  \\
\;\;\textit{where\_clause}\;\;\;\;\;\textit{expr}, \texttt{","}, \textit{where\_conj} \\
\;\;\textit{where\_conj}\;\;\;\;\;\;\texttt{"AND"}, \textit{where\_clause} \\
\midrule
\textbf{Group by clause}  \\
\;\;\textit{groupby\_clause}\;\;\;\;\texttt{"GROUP BY"}, \textit{group\_clause} \\
\;\;\textit{group\_clause}\;\;\;\;\;\textit{expr}, \textit{group\_clause} \\
\midrule
\textbf{Expressions}  \\
\;\;\textit{expr}\;\;\;\;\;\;\;\;\;\textit{value}, \texttt{"BETWEEN"}, \textit{value}, \texttt{"AND"}, \textit{value} \\
\;\;\textit{value}\;\;\;\;\;\;\;\;\textit{col\_ref} \\
\midrule
\textbf{Terminal rules}         \\
\;\;\textit{table\_name}\;\;\;\;\;\;\texttt{"FLIGHTalias0"}       \\
\;\;\textit{column\_name}\;\;\;\;\;\texttt{"FLIGHT\_ID"}       \\
\;\;\textit{col\_ref}\;\;\;\;\;\;\;\texttt{"FLIGHTalias0.FLIGHT\_ID"}      \\
\;\;\textit{col\_alias}\;\;\;\;\;\;\texttt{"DERIVED\_FIELDalias0"}     \\
\;\;\textit{subq\_alias}\;\;\;\;\; \texttt{"DERIVED\_TABLEalias0"}     \\
\bottomrule
\end{tabular}
\caption{Examples for different types of SQL grammar rules. Non-terminal and terminal expressions (in quotation marks) are derived from a non-terminal (left hand side). }
\label{tab:grammars}
\end{table*}

\section{Training}
We implement and train our models using AllenNLP with PyTorch as backend, and conduct experiments on 2 machines each with 4 NVIDIA GeForce GTX $1080$ GPUs and $16$ Intel(R) Xeon(R) CPU E$5-1660$ v$4$ CPUs. The OS is Ubuntu $18.04$ LTS. Averaged running time per model are detailed in ~Table \ref{tab:runtime}.

\paragraph{SQL hyper-parameters} We use Adam optimizer with learning rate selected from $\{0.001, 0.0001\}$. Batch size is selected from $\{1, 4\}$, and we use patience of $15$ epochs. We use EM on the development set as a metric for early stopping and selecting the best hyper-parameters. 
For all models, we use pre-trained GloVe embeddings of size 100, and the target embedding dimension is $100$.Encoder hidden size is selected from $\{200, 300\}$. Dropout is kept fixed at $p=0.5$. We train each model with five random seeds. We perform a grid-search and use accuracy on the development set for model selection.

\elmo{} and \bert{} representations are concatenated to the trainable 100 dimension GloVe embeddings. For \bert{} we use the top layer of the bert-base-uncased model. \elmo{} and \bert{} based models are trained with Noam learning scheduler, with $800$ $600$, or $400$ warm-up steps. 
For the \attnsup{} and \coverage models, the additional loss term scaling hyper-parameter was tuned using the values $\{0.0, 0.1, 0.5, 1.0, 2.5, 5.0\}$. 
For our best performing models, \seqtoseq{}+\coverage{}+\elmo{}, on all datasets, we used an encoder-decoder hidden size of $300$, with coverage loss parameter $0$. Learning rate was set to $0.0001$ for \atis{}, and $0.001$ for the other datasets.

\paragraph{\drop{} hyper-parameters} Similar to SQL, we perform a grid-search to choose hyper-parameters based on the development set accuracy. We tune the following parameters in the specified range and select a single value for all experiments (denoted by \textbf{bold}): learning rate for Adam optimizer in range $\{\textbf{0.001}, 0.0005\}$, batch-size in $\{4, \textbf{16}, 32, 64\}$, and hidden-size for the encoder-decoder LSTMs in $\{\textbf{100}, 200\}$. Dropout is kept fixed at $p=0.2$, gradient clipping is performed with norm-threshold$=5.0$, beam-size is set to $5$, and training is stopped early if the development set accuracy does not improve for $15$ consecutive epochs.

\section{Development Results}

Table~\ref{tab:drop_dev} contains the development set EM for all models on the \drop{} dataset.
Table~\ref{tab:verbose_results_dev} contains the development set EM for all models on all SQL datasets.

\begin{table}[t]
\centering
\footnotesize
\scalebox{1.0}{
\begin{tabular}{lc}
\toprule
\textbf{Model}         &  iid split \\

\midrule
\seqtoseq{}         &  56.9 \\
\;\;+\elmo          &  59.8 \\
\;\;+\bert          &  54.9 \\
\;\;+\attnsup{}     &  55.9 \\
\;\;\;\;+\elmo      &  62.7 \\
\;\;+\coverage      &  54.9 \\
\;\;\;\;+\elmo      &  65.7 \\
\;\;+\attnspan{}    &  57.8 \\
\;\;\;\;+\elmo      &  59.8 \\
\midrule
\grammar{}          &  60.8 \\
\;\;+\elmo          &  67.6 \\
\;\;+\bert          &  65.7 \\
\;\;+\attnsup{}     &  62.7 \\
\;\;\;\;+\elmo      &  69.6 \\
\bottomrule
\end{tabular}}
\caption{\label{tab:drop_dev} iid development set exact match for all models on the \drop{} dataset. 
We no not create a program-split development set for \drop{}, one containing templates not seen in training or test. Instead, we use the same iid development set to choose the best model for both iid and program split settings. Note that this is a more challenging setting, since the model selection for the program split is also done on the basis of an in-distribution development set.}
\end{table}

\renewcommand{\arraystretch}{1.0}
\begin{table*}[t]
\centering
\footnotesize
\resizebox{\textwidth}{!}{
\begin{tabular}{l|ccc|ccc|ccc|ccc}
\toprule
\multirow{2}{*}{\textbf{Model}}         & \multicolumn{3}{c}{Advising} & \multicolumn{3}{c}{ATIS} & \multicolumn{3}{c}{GeoQuery} & \multicolumn{3}{c}{Scholar}\\
                                        &  iid split & Prog. split & Rel. gap &  iid split & Prog. split & Rel. gap &  iid split & Prog. split & Rel. gap &  iid split & Prog. split & Rel. gap\\
\midrule
\seqtoseq{}            & 92.9 & 9.8 & 89.5 & 76 & 11.5 & 84.9 & 67.6 & 27.5 & 59.3 & 73 & 9.3 & 87.3 \\
\;\;+\elmo               & 94.6 & 15.2 & 83.9 & 76.9 & 17.4 & 77.4 & 69.1 & 38.2 & 44.7 & 75.9 & 12 & 84.2 \\
\;\;+\bert              & 94.1 & 14.1 & 85 & 77.8 & 14.8 & 81 & 71.1 & 31.4 & 55.8 & 77.3 & 7.4 & 90.4 \\
\midrule
\;\;+\attnsup{}          & 92.1 & 18 & 80.5 & 74.4 & 22.9 & 69.2 & 67.2 & 43.5 & 35.3 & 69.5 & 13.2 & 81 \\
\;\;\;\;+\elmo{}             & 92.4 & 21.1 & 77.2 & 75.6 & 23 & 69.6 & 67 & 47.7 & 28.8 & 74.2 & 13.8 & 81.4 \\
\;\;\;\;+\bert{}  & 93 & 18.5 & 80.1 & 75.5 & 20.3 & 73.1 & 68 & 47.5 & 30.1 & 73.7 & 13.6 & 81.5 \\
\;\;+\coverage{}     & 93.2 & 16.9 & 81.9 & 75.6 & 19.7 & 73.9 & 70.7 & 43.5 & 38.5 & 74.2 & 9.5 & 87.2 \\
\;\;\;\;+\elmo{}  & 94.9 & 23.2 & 75.6 & 78.4 & 28.9 & 63.1 & 72.2 & 51.2 & 29.1 & 77.8 & 17.5 & 77.5 \\
\;\;\;\;+\bert{}  & 95.4 & 16.8 & 82.4 & 79 & 26.1 & 67 & 74.2 & 51 & 31.3 & 79.1 & 18.4 & 76.7 \\
\;\;+\attnspan{}         & 92.4 & 13.1 & 85.8 & 75.7 & 12.8 & 83.1 & 64.5 & 32.3 & 49.9 & 73.2 & 8.9 & 87.8\\
\;\;\;\;+\elmo{}            & 94.2 & 10.3 & 89.1 & 77.6 & 19.3 & 75.1 & 65.4 & 40.7 & 37.8 & 73.9 & 11.8 & 84 \\
\;\;\;\;+\bert{}  & 94.6 & 14.9 & 84.2 & 76.6 & 15.1 & 80.3 & 67.6 & 35.6 & 47.3 & 75 & 13.6 & 81.9 \\
\midrule
\grammar{}             & 91.1 & 22.5 & 75.3 & 70.1 & 13 & 81.5 & 63.5 & 24.8 & 60.9 & 65.4 & 14.6 & 77.7 \\
\;\;+\elmo               & 91.4 & 15.6 & 82.9 & 60.8 & 13.5 & 77.8 & 58.1 & 21.1 & 63.7 & 66.3 & 14.4 & 78.3 \\
\;\;+\bert               & 93.9 & 17.9 & 80.9 & 61.7 & 5.3 & 91.4 & 64.3 & 19.8 & 69.2 & 66.8 & 13.6 & 79.6 \\
\midrule
\;\;+\attnsup{}         & 91 & 23.5 & 74.2 & 65.9 & 23.6 & 64.2 & 66.6 & 26.8 & 59.8 & 65.2 & 14.8 & 77.3 \\
\;\;\;\;+\elmo             & 91 & 19.4 & 78.7 & 67 & 14.6 & 78.2 & 65.6 & 22.2 & 66.2 & 63.8 & 15.1 & 76.3 \\
\;\;\;\;+\bert  & 91.2 & 14.9 & 83.7 & 59.7 & 3.5 & 94.1 & 65 & 19.1 & 70.6 & 64 & 12.4 & 80.6 \\
\bottomrule
\end{tabular}
}
\caption{\label{tab:verbose_results_dev} Dev EM for all models and all SQL datasets.}
\end{table*}

\end{document}